\documentclass[a4paper, 10pt, conference]{ieeeconf}      

\IEEEoverridecommandlockouts                              
\usepackage{fancyhdr}

\usepackage{graphics} 
\usepackage{balance}
\usepackage{epsfig} 

\usepackage{float} 
\usepackage{pifont}
\usepackage{newunicodechar}
\newunicodechar{✓}{\ding{51}}
\newunicodechar{✗}{\ding{55}}
\usepackage{mathtools,xparse}
\usepackage{amsmath}
\usepackage{amssymb}
\usepackage{amsfonts}
\usepackage{bbm}
\usepackage{algorithm}
\usepackage{algpseudocode}
\usepackage{colortbl}
\usepackage{multirow}
\usepackage{arydshln}
\usepackage{makecell}
\usepackage{algorithm}
\usepackage{hyperref}
\usepackage{booktabs}
\usepackage{subcaption}

\usepackage{xcolor}
\usepackage{soul}

\def\FGPaperID{269} 

\title{\LARGE \bf Disentangled Source-Free Personalization for Facial Expression Recognition with Neutral Target Data}

\begin{document}

\thispagestyle{empty}
\pagestyle{empty}

\author{\parbox{16cm}{\centering
    {\large Masoumeh Sharafi$^1$, Emma Ollivier$^1$, Muhammad Osama Zeeshan$^1$, Soufiane Belharbi$^1$, Alessandro Lameiras Koerich$^2$, Marco Pedersoli$^1$, Simon Bacon$^3$, Eric Granger$^1$}\\
    {\normalsize
    $^1$LIVIA, Dept. of Systems Engineering, ETS Montreal, Canada\\
    $^2$LIVIA, Dept. of Software and IT Engineering, ETS Montreal, Canada\\
    $^3$Dept. of Health, Kinesiology \& Applied Physiology, Concordia University, Montreal, Canada}\\
    {\small \{masoumeh.sharafi.1, emma.ollivier.1, muhammad-osama.zeeshan.1\}@ens.etsmtl.ca}\\
    {\small \{soufiane.belharbi, alessandro.lameiraskoerich, marco.pedersoli, eric.granger\}@etsmtl.ca}\\
    {\small simon.bacon@concordia.ca}
}}


\maketitle

\thispagestyle{fancy}  

\begin{abstract}
Facial Expression Recognition (FER) from videos is a crucial task in various application areas, such as human-computer interaction and health diagnosis and monitoring (e.g., assessing pain and depression). 
Beyond the challenges of recognizing subtle emotional or health states, the effectiveness of deep FER models is often hindered by the considerable inter-subject variability in expressions.  
Source-free (unsupervised) domain adaptation (SFDA) methods may be employed to adapt a pre-trained source model using only unlabeled target domain data,  
thereby avoiding data privacy, storage, and transmission issues. Typically, SFDA methods adapt to a target domain dataset corresponding to an entire population and assume it includes data from all recognition classes. However, collecting such comprehensive target data can be difficult or even impossible for FER in healthcare applications. 
In many real-world scenarios, it may be feasible to collect a short neutral control video (which displays only neutral expressions) from target subjects before deployment. These videos can be used to adapt a model to better handle the variability of expressions among subjects. This paper introduces the Disentangled SFDA (DSFDA) method to address the challenge posed by adapting models with missing target expression data. DSFDA leverages data from a neutral target control video for end-to-end generation and adaptation of target data with missing non-neutral data. Our method learns to disentangle features related to expressions and identity 
while generating the missing non-neutral expression data for the target subject, thereby enhancing model accuracy. Additionally, our self-supervision strategy improves model adaptation by reconstructing target images that maintain the same identity and source expression.  
Experimental results\footnote{\textit{} \href {https://github.com/MasoumehSharafi/DSFDA/}{https://github.com/MasoumehSharafi/DSFDA/}} on the challenging BioVid, UNBC-McMaster and StressID datasets indicate that our DSFDA approach can outperform state-of-the-art adaptation methods.  
\end{abstract}

\section{INTRODUCTION}

Facial expression recognition (FER) has attracted much attention in computer vision due to the breadth of its applications, such as pain and depression assessment, stress estimation, and driver fatigue monitoring. Although state-of-the-art deep FER models \cite{mohan2021fer, ruan2021feature, she2021dive, zhou2022discriminative} have made significant progress, developing applications and models capable of characterizing expression remains a challenge. Deep learning models \cite{ref10}, \cite{ref9}, \cite{ref12} have shown promising performance on various FER tasks. However, their performance can degrade significantly when there is a shift in data distributions between the source (lab) and target (operational) domains. Various unsupervised adaptation (UDA) methods have been

 \begin{figure}[!t]
    \centering
    \includegraphics[width=0.48\textwidth, height=0.62\textwidth]{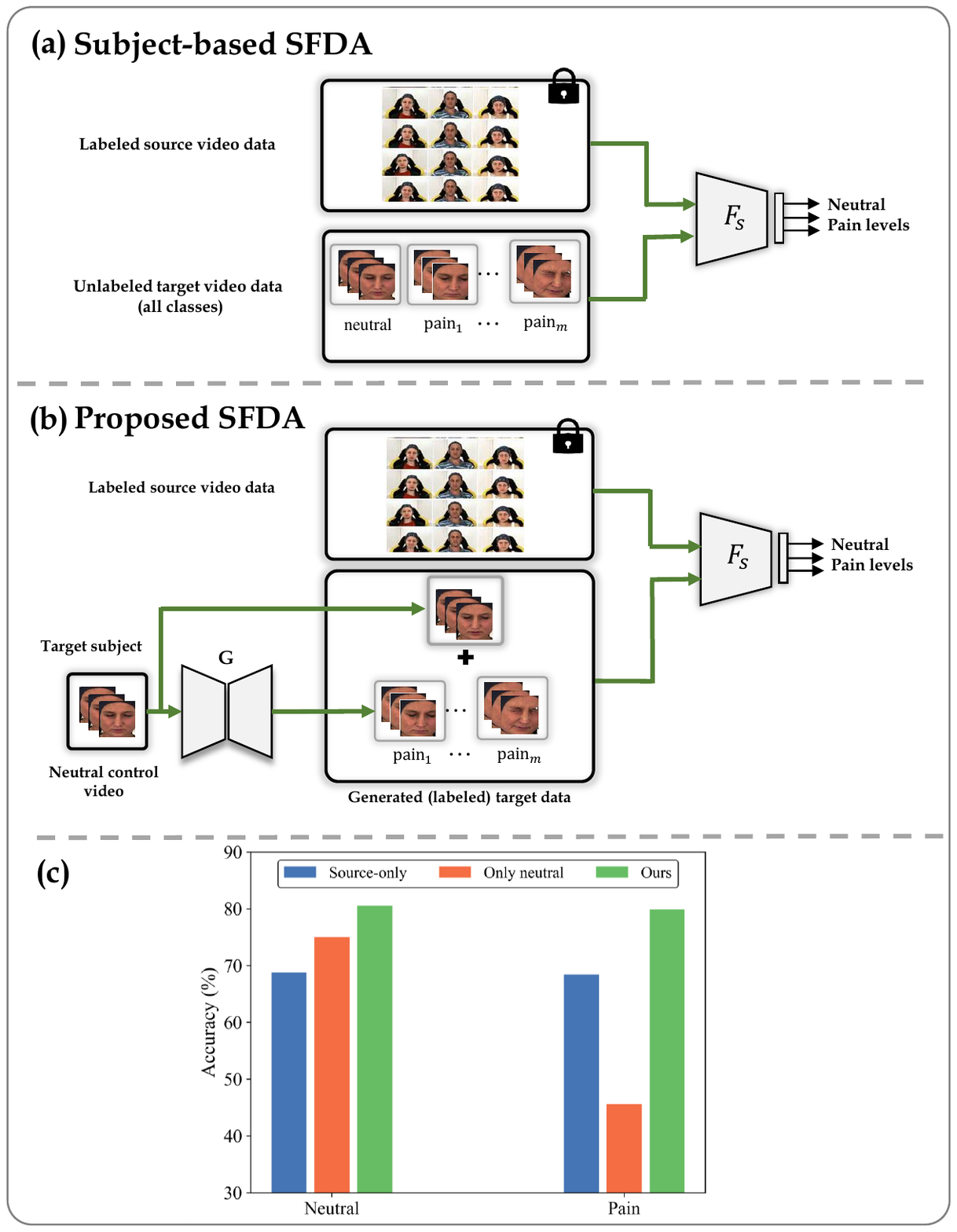}
       \caption{Illustration of SFDA settings using pain estimation as an example. (a) A standard subject-based SFDA setting where the source pre-trained model $F_S$ is adapted through a pseudo-labeling approach using unlabeled target video data from all classes (neutral and pain levels). (b) Our proposed subject-based disentangled SFDA (DSFDA) setting with neutral control target data. The generator $G$ is employed to synthesize the missing classes in the target domain based on the neutral control video. (c) Comparison of accuracy on BioVid data \cite{walter2013biovid} of a source-only pre-trained Inception-V3 CNN ($F_s$ before adaptation), after adaptation with only neutral control data, and with our DSFDA method on neutral and generated pain data. (Lock symbols denote that labeled source data is unavailable during adaptation.)}
    \label{fig:setting}
\end{figure}

\noindent proposed to address the distribution shifts between the source and the target domain \cite{li2018deep, liu2022deep, zeeshan2024subject, zhu2016discriminative}. For instance, the work by Zeeshan et al.~\cite{zeeshan2024subject} introduces a subject-based multi-source domain adaptation (MSDA) method specifically tailored for FER tasks, where each subject is considered as an individual domain for fine-grained adaptation to the variability of expressions among subjects.

State-of-the-art UDA methods \cite{li2018deep, zhu2016discriminative, chen2021cross, ji2019cross, li2020deeper} rely on the availability of the labeled source data during adaptation, which is impractical in real-world scenarios due to concerns for data privacy, data storage, transmission cost, and computation burden. Therefore, to reduce the reliance on source data during the adaptation process, several SFDA methods have been proposed \cite{fang2022source, li2024comprehensive, liang2020we, yang2021exploiting, zhang2023source}. 

SFDA methods typically focus on the scenario where unlabeled target data from all classes is available for the adaptation process (see Fig.~\ref{fig:setting}(a)). 
However, in real-world video-based health applications, person-specific data representing non-neutral expressions is typically costly or unavailable. A short neutral control video may, however, be collected for target individuals and used to personalize a model to the variability of an individual's diverse expressions. In practice, collecting neutral target data for adaptation is generally easier than gathering non-neutral emotional data. Annotating neutral expressions is more straightforward and less subjective than interpreting emotional expressions, which can vary considerably among expert annotators. Creating scenarios to elicit specific emotions can be complex and time-consuming, and may raise ethical and privacy concerns.  
In contrast, capturing neutral expressions is less intrusive and may encourage more willing participation, as it avoids the need to act out emotional states.   

This paper introduces a new setting for SFDA, where data from all classes is available for source domain subjects to pre-train a deep FER model, while only neutral control data is available for target domain subjects to adapt this model. In the standard SFDA setting, data from all classes is available for the source and target domains. In contrast, our proposed setting assumes that only neutral data is available for adaptation. At inference time, however, the FER model will process data from all classes. Fig.~\ref{fig:setting}(c) shows the limitation of using a source-only pre-trained deep FER model without adaptation (in blue) or when fine-tuning with only neutral control target data (in orange). The accuracy of models is compared across neutral and pain classes (all pain levels combined). A model adapted only to neutral data becomes biased toward the neutral class and struggles to generalize to non-neutral expression. This is shown by its significantly lower accuracy on pain expressions compared to the source-only model.

In this paper, a new personalized SFDA method is proposed for FER with our new setting (see Fig.~\ref{fig:setting}(b)). The Disentangled SFDA (DSFDA) method is proposed to address the challenges of personalized SFDA using neutral control target data by generating the missing non-neutral expressive image data for target subjects. Using a dual-generator architecture pre-trained on source subjects, a neutral image is transformed into one that exhibits the missing non-neutral expression. By combining features from two input images, the method synthesizes expressive images that belong to the missing non-neural target expressions, enabling accurate FER predictions for target subjects even without explicit data for those expressions. Our approach is further enhanced using self-supervised learning, which ensures the reconstruction of input target images with the same identity and source expression. This process improves the model's adaptability and predictive accuracy in the absence of source data. As shown in Fig.~\ref{fig:setting}(c), DSFDA overcomes the challenges of domain shifts and incomplete target data, ensuring a high level of performance across diverse target domains. 

Our contributions can be summarized as follows. 
    \noindent \textbf{(1)} A novel SFDA setting is introduced, where all classes in the source data are utilized to pre-train the source model. For target adaptation, however, only neutral control target data is available to adapt the model for a target subject.
    \noindent \textbf{(2)} We propose the DSFDA method, a one-stage approach tailored to our novel setting, that combines data generation and target domain adaptation to simultaneously generate non-neutral target image data and adapt a FER model.
    \noindent \textbf{(3)} An extensive set of experimental results indicating that a FER model adapted using our DSFDA method can outperform state-of-the-art SFDA methods on the BioVid, UNBC-McMaster, and StressID datasets. The method scales well to the number of subjects and is robust across various subjects, target domains, and datasets.

\section{Related Work}

\subsection{Source-Free Domain Adaptation}
UDA methodologies \cite{zhu2016discriminative, chen2021cross, ji2019cross, li2020deeper} require simultaneous access to both labeled source domain and unlabeled target domain data during training. Some facial analysis datasets are characterized by concerns about privacy and sensitivity. SFDA methods have been proposed to address the challenges of privacy and memory storage. SOTA methods SFDA can be classified into two distinct groups \cite{fang2022source}: data-based methods \cite{ding2022source, hong2022source, hou2020source,kurmi2021domain, tian2021source, yang2022source} and the model-based methods \cite{guichemerre2024source, liang2020we, yang2021transformer}. 

\vspace{+0.1cm}
\noindent \textbf{Data-based methods:} \cite{hou2020source} used BN statistics to generate source-like images, which transfer the style by aligning batch-wise feature statistics of generated source-like image features to that stored in the BN layer. Instead of directly generating source-like data, other studies focus on approximating and generating the underlying data distribution. For example, \cite{ding2022source} utilized source classifier weights and target pseudo-labels computed through spherical $k$-means clustering to estimate the source feature distribution. Next, the proxy source data can be extracted from the approximated source distribution, and a conventional DA approach is used to achieve cross-domain alignment.

\vspace{+0.1cm}
\noindent \textbf{Model-based methods:} Conti et al.~\cite{conti2022cluster} introduced a cluster purity score to obtain reliable pseudo-labels using $k$-means clustering. The target feature extractor is then trained using two augmentations of input images instead of raw input data. The reduced target dataset is trained using the reliable pseudo-labels obtained with the self-supervised feature extractor. Guo et al. \cite{guo2023ltval} incorporate Tsallis entropy into the information maximization constraint to enhance the confidence of prediction labels while reducing confidence bias. 

Unlike the promising results of SFDA methods in benchmark classification tasks, current SFDA approaches are typically designed under the assumption of an equal number of classes in both the source and target domains. However, in FER tasks, non-neutral expression data (missing classes) may be unavailable, with only a short neutral video available for adaptation. This leads to a label shift between the target adaptation and testing phases, ultimately degrading classification performance. To mitigate this shift between the source and target domains, generative models are utilized to synthesize the missing classes within the target domain. These models aim to produce high-quality images or videos that accurately represent the emotional expressions absent in the original target dataset. In the literature, a variety of techniques have been explored for this purpose, ranging from traditional generative adversarial networks (GANs) and variational autoencoders (VAEs) to more advanced frameworks such as conditional GANs, diffusion models, and disentanglement methods.

\subsection{Facial Expression Generation and Disentanglement}
Generative modeling, a branch of machine learning (ML), concentrates on understanding the underlying distribution of training data to enable the generation of new samples that mimic the statistical features of that data. Over time, significant progress has been made in this field. Most common techniques like GANs \cite{arjovsky2017wasserstein, azari2024emostyle, brock2018large, choi2020stargan, zhang2017age}, VAEs \cite{kingma2013auto, razavi2019generating, vahdat2020nvae}, and diffusion models \cite{dhariwal2021diffusion, kim2023dcface, nichol2021improved, rombach2022high} have significantly improved the quality and variety of the samples generated. For facial expression generation, \cite{siddiqui2022fexgan} developed FExGAN-Meta for meta-humans, while ImaGINator \cite{wang2020imaginator} employed CGANs for creating facial expression video sequences. Niinuma et al.~\cite{niinuma2022facial} used a GAN-based approach to synthesize facial expressions for AU intensity estimation, improving performance on imbalanced and domain-shifted datasets through personalized networks. Recent works have emphasized domain generalization and adaptation for cross-corpus facial expression recognition. Yin et al.~\cite{yin2021contrastive} introduced the FATE model that uses contrastive learning and facial animation warping to achieve superior recognition of emotions in the cross-domain. Furthermore, DeepFN \cite{hernandez2022deepfn} proposed deep face normalization for robust recognition of action units between individuals. Kim et al.~\cite{yang2019learning} leveraged single-image temporal cues for action unit detection, enhancing the understanding of subtle expressions.

In ML, disentanglement refers to the process of learning representations where distinct factors of variation in the data are captured in separate, interpretable components \cite{def}. This property has proven beneficial in applications such as data augmentation, image synthesis, and improving model generalization. Several disentanglement methods have been proposed in the literature, and Wang et al.~\cite{ref15} suggest classifying these techniques according to various criteria. Evaluating disentanglement, however, remains a challenging task, as there is no universal metric capturing all aspects of factor separation. A recent survey provides a detailed review of existing disentanglement metrics, highlighting their limitations and suitability across use cases \cite{carbonneau2022measuring}. Disentanglement is particularly relevant in facial expression recognition (FER), where expression features are often entangled with identity-specific traits. Generative adversarial networks (GANs) are frequently employed to decouple expression and identity features, thereby enhancing both model performance and interpretability \cite{ref12, ref11, dis_gan}. For instance, the IPD-FER \cite{ref2} method disentangles identity, pose, and expression components, while Halawa et al. \cite{ref14} propose a GAN-based approach that separates expression features without requiring identity labels.

Beyond GANs, other disentanglement techniques include VAE methods \cite{ref18}. Li et al.~\cite{ref16} use the Hilbert-Schmidt independence criterion to separate expression from identity features, showcasing mathematical criteria in disentanglement. Mutual information minimization is another strategy for this goal \cite{ref17}. These studies illustrate various methods to achieve robust and interpretable disentangled representations in facial expression recognition.

Given only neutral videos in the target domain, we generate non-neutral expressions to adapt a source model. This two-stage approach re-frames the SFDA task by using a generator (e.g., VAE, GAN, or DM) trained on the source domain to synthesize expressive frames. These are combined with neutral target frames to adapt the model. However, current generators struggle to capture subtle pain intensities as a result of the complexity of nuanced expressions.

SOTA SFDA methods assume consistent classes across adaptation and testing, but FER tasks often lack non-neutral target expressions due to data collection challenges. This leads to label and domain shifts that affect performance. Generative models try to fill the gap but struggle with subtle expressions such as pain. Our DSFDA method overcomes this by disentangling expression from identity, increasing recognition accuracy.

\begin{figure*}[t!]
    \centering
    \includegraphics[width=1\textwidth, height=0.47\textwidth]{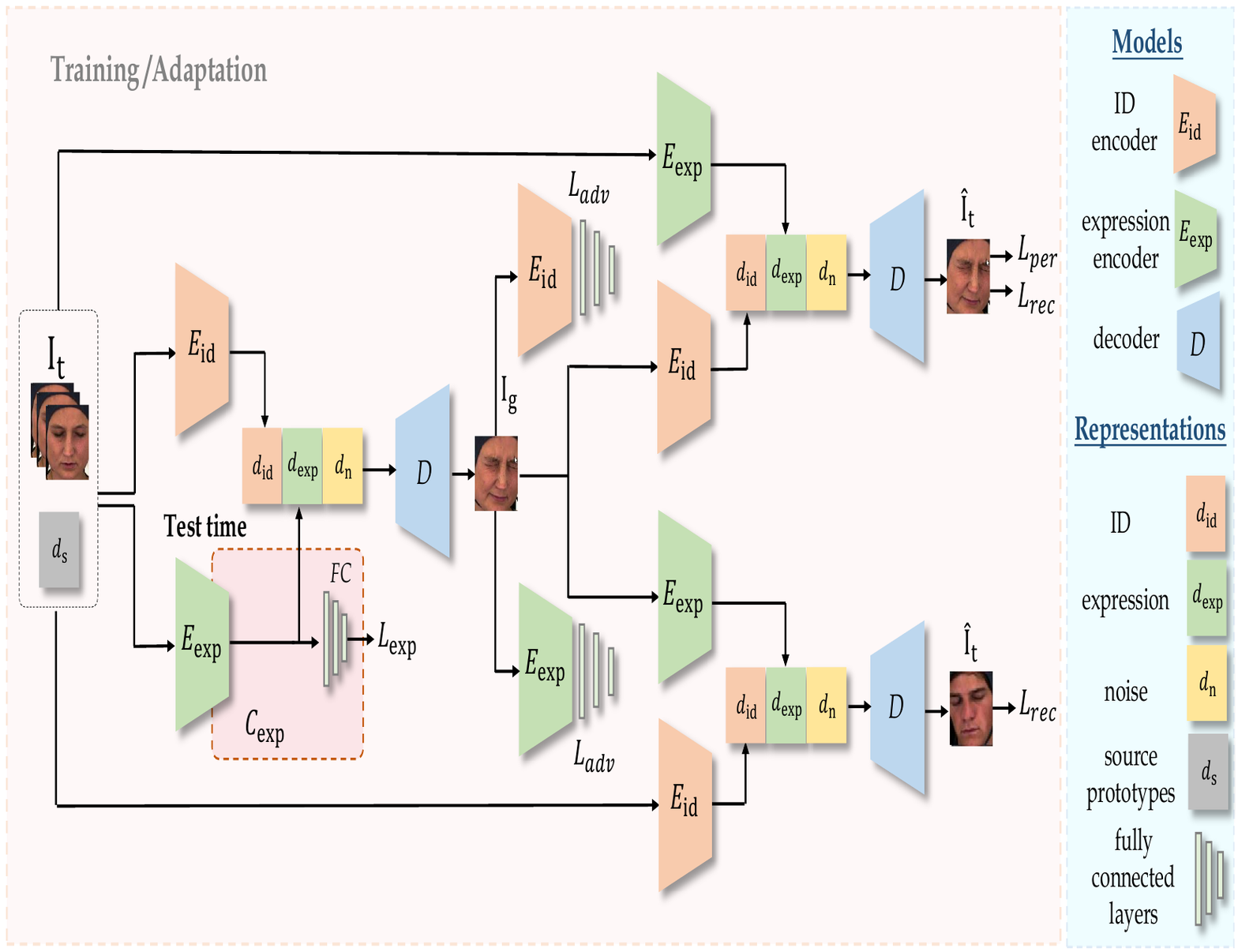}
    \caption{Training and testing architecture of our DSFDA method. In the training and adaptation part (\textit{pink}), the identity and expression features of input images are used to generate images with the identity of the target image and the expression of the source prototype embeddings through adversarial training. The model learns to disentangle the expression and identity information through generation. At test time (\textit{red}), the expression encoder with fully connected layers is used to classify expressions.}
    \label{fig:dis-arc}
\end{figure*}

\section{Proposed Method}

Fig.~\ref{fig:dis-arc} shows the training and testing architecture of our Disentangled SFDA (DSFDA) method for adapting a FER model using only neutral control video per target subject. To produce the non-neutral expressions for the target data and align with the source domain, a disentanglement method is used to simultaneously generate and adapt the target data by reducing the perceptual and reconstruction loss between the input image and the generated image.
 
\noindent \textbf{Notation}. In the proposed method, $\mathbf{I}_{\text{t}}$ represents the target image, which can either be an expression image $\mathbf{I}_{\text{exp}}$ with its corresponding one-hot label $y_{\text{exp}}$ or an identity image $\mathbf{I}_{\text{id}}$ with its corresponding one-hot label $y_{\text{id}}$. The image generated by the model is represented as \(\mathbf{I}_{\text{g}}\), while \(y_{\text{g}}\) is the one-hot label of the generated image. The number of identity classes is denoted as \(c_{\text{id}}\), and the number of expression classes is represented as \(c_{\text{exp}}\). The distribution of identity images is indicated by \(p_{\text{id}}\), the distribution of expression images by \(p_{\text{exp}}\), and the distribution of generated images by \(p_{\text{g}}\). The reconstructed target image is denoted as \(\hat{\mathbf{I}}_{\text{t}}\), may represent either the reconstructed identity image \(\hat{\mathbf{I}}_{\text{id}}\) or the reconstructed expression image \(\hat{\mathbf{I}}_{\text{exp}}\). The identity representation extracted by the identity encoder is denoted as $\mathbf{d}_{\text{id}}$, while $\mathbf{d}_{\text{exp}}$ represents the expression embedding of the expression image. The source domain prototypes are indicated by $\mathbf{d}_{\text{s}}$. Additionally, \(n\) represents the number of training samples, and \(\hat{y}_{i}^{j}\) is the \(j\)-th output value from the softmax classifier for the \(i\)-th sample.

\subsection{Model Architecture}

DSFDA builds upon the architecture proposed in \cite{ref12}, where facial expression information is disentangled from identity to enhance expression classification and domain adaptation. The model consists of a generator and two discriminators: the generator includes two encoders—an identity encoder ($E_{\text{id}}$) and an expression encoder ($E_{\text{exp}}$)—as well as a shared decoder ($D$). The discriminators, $D_{\text{id}}$ and $D_{\text{exp}}$, are constructed using the respective encoders followed by fully connected layers to perform identity and expression classification.

Given a pair of input images, $\mathbf{I}_{\text{id}}$ and $\mathbf{I}_{\text{exp}}$, the generator synthesizes an image $\mathbf{I}_{\text{g}}$ that combines the identity from $\mathbf{I}_{\text{id}}$ and the expression from $\mathbf{I}_{\text{exp}}$. This generated image is then assessed by the two discriminators to verify the preservation of the intended identity and expression. To ensure faithful disentanglement and reconstruction, the generator is further used to regenerate $\hat{\mathbf{I}}_{\text{id}}$ and $\hat{\mathbf{I}}_{\text{exp}}$ from $\mathbf{I}_{\text{g}}$, aiming to recover the original identity and expression inputs. 
Training is guided by a combination of reconstruction, perceptual, and adversarial losses, along with an expression classifier $C_{\text{exp}}$ trained on real data to support accurate expression recognition.

\subsection{Objective Functions}

\noindent \textbf{(1) Classifier:}
Fully connected layers of the classifier $C_{\text{exp}}$ are trained using real images $\mathbf{I}_{\text{exp}}$ with cross-entropy loss:
\begin{equation}
\mathcal{L}_{\text{exp}} =  - \mathbb{E}_{(\mathbf{I}_{\text{exp}},y_{\text{exp}}) \sim p_{\text{exp}}} \sum_{e=1}^{c_{\text{exp}}}\mathbbm{1}[y_{\text{exp}}=e]\log(C_{\text{exp}}(\mathbf{I}_{\text{exp}}))  
\end{equation}

\noindent \textbf{(2) Discriminator:}
To operate the adversarial method, two discriminators are trained with real images $\mathbf{I}_{\text{id}}$ and $\mathbf{I}_{\text{exp}}$ to recognize either identity or pain expression. We employ an expression discriminator $D_{\text{exp}}$ with $c_{\text{exp}}$ classes. Similarly, an identity discriminator $D_{\text{id}}$, configured with $c_{\text{id}} + 1$ classes, guides the generator to produce realistic images with the specified identities. The loss functions for the identity discriminator $\mathcal{L}_{\text{dis\_id}}$ and the expression discriminator $\mathcal{L}_{\text{dis\_exp}}$ are defined by:
\begin{equation}
\begin{split}
\mathcal{L}_{\text{dis\_id}} =  - \mathbb{E}_{(\mathbf{I}_{\text{id}},y_{\text{id}}) \sim p_{\text{id}}} \sum_{i=1}^{c_{\text{id}+1}}\mathbbm{1}[y_{\text{id}}=i]\log(D_{\text{id}}(\mathbf{I}_{\text{id}})) \\ - \mathbb{E}_{(\mathbf{I}_{\text{g}},y_{\text{g}}) \sim p_{\text{g}}} \sum_{g=1}^{c_{\text{g}}}\mathbbm{1}[y_{\text{g}}=g]\log(D_{\text{g}}(\mathbf{I}_{\text{id}}))
\end{split}
\end{equation}
\begin{equation}
\begin{split}
\mathcal{L}_{\text{dis\_exp}} = - \mathbb{E}_{(\mathbf{I}_{\text{exp}},y_{\text{exp}})\sim p_{\text{exp}}} \sum_{e=1}^{c_{\text{exp}}}\mathbbm{1}[y_{\text{exp}}=e]\log(D_{\text{\text{exp}}}(\mathbf{I}_{\text{exp}}))  
\end{split}
\end{equation}
\noindent where $y_{\text{g}}$ is the label of the generated image, $p_{\text{id}}$, $p_{\text{exp}}$, and $p_{\text{g}}$ represent the distributions of identity images, expression images, and generated images, respectively, and $c_{\text{id}}$, $c_{\text{exp}}$, and $c_{\text{g}}$ are the number of identity, expression, and generated image classes.

Given that the generated images should replicate the overall appearance of the identity input image, an additional class for real and fake images is included in the identity discriminator. Consequently, only the identity discriminator processes the generated images. The expression classifier’s role is to classify the expressions of the images. Using real expression data for the expression discriminator during training is therefore advantageous. Using generated images with incorrect emotions can destabilize the training process.

\noindent \textbf{(3) Generator:}
The generator consists of a decoder and two encoders: one for expression and the other for identity. During training, all components of the generator are updated simultaneously. A reconstruction loss is employed to ensure the generation of accurate images, where the reconstructed identity image $\hat{\mathbf{I}}_{\text{id}}$ closely resembles the input identity image $\mathbf{I}_{\text{id}}$, and the reconstructed expression image $\hat{\mathbf{I}}_{\text{exp}}$ closely matches the input expression image $\mathbf{I}_{\text{exp}}$: 
\begin{equation}
\mathcal{L}_{\text{rec}} = \mathbb{E}_{\mathbf{I}_{\text{id}}\sim p_{\text{id}}} \lVert \hat{\mathbf{I}}_{\text{id}} - \mathbf{I}_{\text{id}} \rVert_{1} + \mathbb{E}_{\mathbf{I}_{\text{exp}}\sim p_{\text{exp}}} \lVert \hat{\mathbf{I}}_{\text{exp}} - \mathbf{I}_{\text{exp}} \rVert_{1}
\end{equation}
In addition, perceptual loss, defined by MSE loss, is used to ensure that identity features produced by the discriminator are the same for real and generated images, $\mathbf{I}_{\text{id}}$ and $\mathbf{I}_{\text{g}}$: 
\begin{equation}
\mathcal{L}_{\text{per}} = \mathbb{E}_{\mathbf{I}_{\text{id}}\sim p_{\text{id}}} \lVert \mathbf{d}_{\text{id}}(\mathbf{I}_{\text{g}}) - \mathbf{d}_{\text{id}}(\mathbf{I}_{\text{id}}) \rVert_{2}
\end{equation}
\noindent where $\mathbf{d}_{\text{id}}(.)$ denotes features extracted by the discriminator specialized in identity recognition.

Discriminators $D_{\text{id}}$ and $D_{\text{exp}}$ are trained to recognize identity and expression. To make the generator fool both discriminators, the following adversarial loss is applied:
\begin{equation}
\begin{split}
\mathcal{L}_{\text{adv}} =  - \lambda_\text{adv1}\mathbb{E}_{(\mathbf{I}_{\text{id}},y_{\text{id}})} \sum_{i=1}^{c_{\text{id}}}\mathbbm{1}[y_{\text{id}}=i]\log(D_{\text{id}}(\mathbf{I}_{\text{d}}))  \\ - \lambda_\text{adv2}\mathbb{E}_{(\mathbf{I}_{\text{exp}},y_{\text{exp}})} \sum_{e=1}^{c_{\text{exp}}}\mathbbm{1}[y_{\text{exp}}=e]\log(D_{\text{exp}}(\mathbf{I}_{\text{g}}))
\end{split}
\end{equation}
\noindent where $D_{\text{id}}(.)$ and $D_{\text{exp}}(.)$ are, respectively, the predicted class distribution by identity and expression discriminators, and $y_{\text{exp}}$ and $y_{\text{id}}$ denote labels of expression and identity for real images.

The total loss function for the generator is:  
\begin{equation}
\mathcal{L}_{\text{gen}} = \lambda_\text{rec}\mathcal{L}_{\text{rec}} + \lambda_\text{per}\mathcal{L}_{\text{per}} + \mathcal{L}_{\text{adv}}
\end{equation}
In Eqs.~(6) and (7), $\lambda_{\text{adv1}}$, $\lambda_{\text{adv2}}$, $\lambda_{\text{rec}}$, and
$\lambda_{\text{per}}$ are hyperparameters used to adjust the impact of individual losses.

\section{Results and Discussion}

\subsection{Experimental Methodology}

\noindent\textbf{Datasets.} We evaluate the performance of our model using the BioVid Heat Pain dataset \cite{walter2013biovid} and the UNBC-McMaster Shoulder Pain dataset \cite{lucey2011painful}.

The BioVid dataset is divided into five distinct parts, each varying in terms of subjects, labeling, modalities, and tasks. In our experimentation, we use Part A of the dataset, which includes data from 87 subjects, encompassing four pain levels (PA1, PA2, PA3, and PA4) in addition to a neutral (BL1). The modalities in Part A consist of frontal video recordings and biological signals. In this study, we focus on the highest pain level and the neutral state, using only the frames extracted from frontal video recordings. Each subject has 20 videos per class, each lasting 5.5 seconds. According to \cite{werner2017analysis}, the PA4 dataset does not show any facial activity in the first two seconds of the videos and has demonstrated that initial pain intensities do not elicit any facial activities. They suggest focusing solely on 'no pain' and the highest pain intensities. Therefore, we only consider frames after the first two seconds to exclude the initial part of the sequence that did not indicate any response to pain. 

The UNBC-McMaster dataset comprises 200 pain videos from 25 subjects. Each video frame is annotated with a pain intensity score based on the PSPI scale in a range of 0 to 15. To address the significant imbalance in pain intensity levels, we adopt the quantization strategy proposed in \cite{rajasekhar2021deep}, categorizing pain intensities into five discrete levels: 0 (neutral), 1 (1), 2 (2), 3 (3), 4 (4–5), and 5 (6–15).

The StressID dataset \cite{chaptoukaev2023stressid} comprises facial video recordings from 54 participants, resulting in approximately 918 minutes of annotated visual data. In this work, we focus solely on the visual modality. Each frame is labeled as either neutral (0) or stress (1), based on the participant’s self-reported stress level. Specifically, labels are assigned using a binary scheme where recordings with a stress score below 5 are marked as neutral, and those with a score of 5 or higher are marked as stressed.

\noindent\textbf{Experimental Protocol.} In our experiments, each dataset is divided into two domains: source and target. The \textit{first experiment} is conducted on the BioVid dataset, where 77 subjects form the source domain and the remaining 10 subjects constitute the target domain, with each target subject treated independently. The evaluation follows a three-phase protocol: source training, target adaptation, and target testing. During the source training phase, the model is trained on data from the 77 source subjects, using both neutral and pain level 4 images. In the adaptation phase, the model is adapted to the target data, which consists of neutral videos from the 10 target subjects. Finally, in the target testing phase, the model is evaluated using 12 videos per target subject, covering both neutral and pain expressions. In the \textit{second experiment}, we use the UNBC-McMaster dataset, designating 19 subjects as the source domain and the remaining 5 as the target. A similar protocol is applied. The \textit{third experiment} uses the StressID dataset, where 44 subjects are assigned to the source domain and 10 subjects to the target domain. To further evaluate the generalization capabilities of our method, we conduct a \textit{cross-dataset experiment}, where the source domain comprises 20 subjects from the UNBC-McMaster dataset, and the target domain includes 10 subjects from BioVid.

\begin{table*}[t!]
\centering
\renewcommand{\arraystretch}{1.5} 
\caption{Accuracy (\%) of the proposed method and SOTA methods on the BioVid dataset for 10 target subjects with all 77 sources. Bold text shows the highest accuracy. Note: SFDA (77-shot) indicates that 77 prototypes were used from the source data.}\label{tab:SFDA-DE_result.}\label{tab:SFDA-DE_result}
\resizebox{1 \textwidth}{!}{%
	\begin{tabular}{c|c||cccccccccc|c}
		\hline
		\textbf{Setting} & \textbf{Methods} & \textbf{Sub-1} & \textbf{Sub-2} & \textbf{Sub-3} & \textbf{Sub-4} & \textbf{Sub-5} & \textbf{Sub-6} & \textbf{Sub-7} & \textbf{Sub-8} & \textbf{Sub-9} & \textbf{Sub-10} & \textbf{Avg.} \\
	    \hline
     \hline
         Source-only & Source model & 66.11 & 55.55 & 86.36 & 85.11 & 87.11 & 59.00 & 75.66 & 70.44 & 52.66 & 48.22 & 68.62 \\
        \hline
        \multirow{3}{*}{\makecell{SFDA}}  & SHOT \cite{liang2020we} & 55.78 & 47.76 & 41.05 & 52.44 & 54.67 & 48.89 & 54.46 & 49.22 & 54.86 & 44.44 & 50.35 \\
        & NRC \cite{yang2021exploiting} & 59.33 & 39.38 & 84.00 & 72.89 & 79.67 & 42.67 & 63.92 & 53.95 & 54.89 & 42.47 & 60.31 \\
         & DSFDA (ours) & \textbf{77.00} & \textbf{75.11} & \textbf{90.89} & \textbf{85.67} & \textbf{88.11} & \textbf{67.48} & \textbf{89.22} & \textbf{87.56} & \textbf{69.22} & \textbf{72.11} & \textbf{80.24}\\
        \hline
        \multirow{3}{*}{\makecell{SFDA \\ ($77$-shot)}} & SHOT \cite{liang2020we} & 85.67 &\textbf{75.11} & 90.89 & 85.36& 87.11& 59.11& 76.86& 84.22& 60.33&\textbf{86.22} & 79.09\\
        & NRC \cite{yang2021exploiting} & 82.78 & 71.40& 90.91 & 85.22& 87.56 & 59.31 &83.73 & 85.89& 65.89& 82.11&79.48\\
        & DSFDA (ours) & \textbf{89.56} & \textbf{75.11} & \textbf{91.00} & \textbf{85.56} & \textbf{87.78} & \textbf{68.33} & \textbf{86.78} & \textbf{88.33} & \textbf{72.22} & 74.11 & \textbf{81.88}\\
        \hline
        Oracle & Fine-tune & 97.11 & 91.43 & 96.76 & 97.89 & 96.11 & 92.30 & 90.01 & 95.09 & 98.22 & 97.00 & 95.19\\
        \hline
	\end{tabular}
}
\end{table*}

\noindent\textbf{Experimental Settings.} To validate our approach, several settings are explored: (1) Source-only, where a pre-trained model on the source domain is directly tested on the target domain without any adaptation. It serves as a baseline for evaluating how well the model generalizes to unseen data; (2) SFDA, where only neutral images from the target subjects are used for adaptation, while testing is performed on both neutral and pain images. This setting is introduced to investigate how well the model performs when adapted using only neutral images and to compare its results to a scenario where both neutral and pain images are available for adaptation; (3) SFDA ($k$-shot), where we utilized target neutral data and $k$ prototypes (images) from the source domain. The image prototypes from the source data were selected using $K$-means clustering, with $k$ being the number of subjects in the source domain. Therefore, one prototype was selected for each subject. The goal here is to investigate whether pain prototypes, drawn from the source domain, can effectively guide the model in recognizing pain in the target domain; and (4) Oracle, an upper-bound scenario, achieved by fine-tuning the source model with labeled neutral and pain images from the target data.

\noindent\textbf{Implementation Details.} In our experiments, we used the Inception-V3 CNN with Instance Normalization as the backbone for the encoders. We trained the entire model on the source data using a learning rate of \(10^{-5}\) for 100 epochs. To adapt the model to the target subjects, we trained with a learning rate of \(10^{-4}\) for 25 epochs, employing a batch size of 32. For the expression transfer, we utilized two inputs corresponding to the expression and identity labels. In our experiments, we maintained consistency by using the same dataset for both the identity and expression encoders. For image preprocessing, we resized the input images to \(128 \times 128\) pixels and applied data augmentation techniques, including random cropping, horizontal flipping, and color adjustments during the training phase. Throughout our experiments, we set the hyperparameters empirically as $\lambda_{\text{rec}}=5$, $\lambda_{\text{per}}=1$, $\lambda_{\text{adv1}}=0.2$, and $\lambda_{\text{adv2}}=0.8$. For the $k$-shot SFDA setting, we define $k$ as the number of source subjects used during training. Specifically, we set $k=77$ for the BioVid dataset, $k=19$ for the UNBC-McMaster dataset, and $k=44$ for the StressID dataset. All experiments were conducted on a single NVIDIA A100 GPU with 48 GB of memory. The source model was trained for 100 epochs, taking approximately 18 hours in total. Compared to baseline methods, our approach is computationally more efficient, as it performs generation and adaptation in a single step—eliminating the need to generate images first and then use them separately for adaptation.

\begin{table}[t!]
\centering
\LARGE
\renewcommand{\arraystretch}{1.5} 
\caption{Accuracy (\%) of the proposed method and SOTA methods on the UNBC-McMaster dataset for five target subjects with all 19 sources. Bold text shows the highest accuracy. Note: SFDA (19-shot) indicates that 19 prototypes were used from the source data.}\label{tab:SFDA-DE_mcmaster}
\resizebox{0.47\textwidth}{!}{%
	\begin{tabular}{c|c||ccccc|c}
		\hline
		\textbf{Setting} & \textbf{Methods} & \textbf{Sub-1} & \textbf{Sub-2} & \textbf{Sub-3} & \textbf{Sub-4} & \textbf{Sub-5} & \textbf{Avg.} \\
	   \hline
         \hline
         Source-only & Source model & 68.56 & 78.32 & 60.37 & 65.54 & 33.72 & 61.30 \\
        \hline
        \multirow{3}{*}{\makecell{SFDA}} & SHOT \cite{liang2020we} & 69.93 & 70.84 & 42.24 & 63.56 & 34.41 & 56.19 \\
        & NRC \cite{yang2021exploiting} & 63.33 & 69.13 & 52.24 & 63.28 & 33.24 & 56.24\\
        & DSFDA (ours) & \textbf{78.23} & \textbf{84.39} & \textbf{89.63} & \textbf{78.99} & \textbf{66.56} & \textbf{79.56}\\
        \hline
        \multirow{3}{*}{\makecell{SFDA \\ (19-shot)}} & SHOT \cite{liang2020we} & 77.51 & 79.42 & 42.99 & 65.65 & 36.47 & 60.41 \\
        & NRC \cite{yang2021exploiting} & 77.26 & 78.56 & 64.61 & 77.89 & 43.53 & 68.37\\
        & DSFDA (ours) & \textbf{78.23} & \textbf{89.45} & \textbf{89.63} & \textbf{78.85} & \textbf{65.23} & \textbf{80.27}\\
        \hline
        Oracle
        & Fine-tune & 95.63 & 98.44 & 98.11 & 97.54 & 98.25 &97.59\\
        \hline
	\end{tabular}
}
\end{table}

\begin{table*}[h!]
\centering 
\renewcommand{\arraystretch}{1.5} 
\caption{Accuracy (\%) of the proposed method on the StressID dataset for 10 target subjects with all 44 source subjects. Bold text shows the highest accuracy.}\label{tab:stress}
\resizebox{\textwidth}{!}{%
	\begin{tabular}{c||cccccccccc|c}
		\hline
		\textbf{Method (Source→Target)} & \textbf{Sub-1} & \textbf{Sub-2} & \textbf{Sub-3} & \textbf{Sub-4} & \textbf{Sub-5} & \textbf{Sub-6} & \textbf{Sub-7} & \textbf{Sub-8} & \textbf{Sub-9} & \textbf{Sub-10} & Avg. \\
	    \hline
     \hline
         Source-only & 38.96 & 41.21 & 65.53 & 42.04 & 55.16 & 65.51 & 69.43 & 60.78 & 53.62 & 55.63 & 54.79 \\
        \hline
        DSFDA   & 73.47 & 69.39 &  87.12 & 69.74 & 79.87 & 87.39 & 82.80 & 83.89 & 75.03 & 77.39 & 78.61\\
         DSFDA (44-shot) & \textbf{75.17} & \textbf{71.09} & \textbf{88.82} & \textbf{71.44} & \textbf{81.57} & \textbf{89.09} & \textbf{84.50} & \textbf{85.59} & \textbf{72.73} & \textbf{79.09} & \textbf{79.91}\\
        \hline
	\end{tabular}
}
\end{table*}

\begin{table*}[h!]
\centering
\renewcommand{\arraystretch}{1.5} 
\caption{Accuracy (\%) of the proposed method and state-of-the-art methods on the BioVid dataset for 10 target subjects with all 19 sources from UNBC-McMaster. Bold text shows the highest accuracy. }\label{tab:mc_to_bio}
\resizebox{\textwidth}{!}{%
	\begin{tabular}{c|c||cccccccccc|c}
		\hline
		\textbf{Setting} & \textbf{Method (Source→Target)} & \textbf{Sub-1} & \textbf{Sub-2} & \textbf{Sub-3} & \textbf{Sub-4} & \textbf{Sub-5} & \textbf{Sub-6} & \textbf{Sub-7} & \textbf{Sub-8} & \textbf{Sub-9} & \textbf{Sub-10} & Avg. \\
	    \hline
     \hline
         Source-only & Source model & 50.00 & 50.00 & 50.11 & 50.33 & 57.89 & 52.78 & 58.00 & 79.44 & 65.11 & 48.22 & 56.48 \\
        \hline
        \multirow{3}{*}{\makecell{SFDA}} & SHOT \cite{liang2020we} & 63.33 & 46.48 & 52.22 & 38.11 & 56.67 & 51.33 & 56.13 & 42.22 & 46.75 & 55.33 & 50.86\\
        & NRC \cite{yang2021exploiting} & 50.33 & 51.17 & 50.11 & 45.56 & 56.56 & 51.00 & 57.93 & 50.78 & 51.33 & 58.22 & 52.30 \\
        & DSFDA (ours) & \textbf{71.78} & \textbf{62.56} & \textbf{78.11} & \textbf{74.22} & \textbf{77.33} & \textbf{62.33} & \textbf{90.56} & \textbf{92.33} & \textbf{65.00} & \textbf{65.89} & \textbf{74.01}\\
        \hline
	\end{tabular}
}
\end{table*}

\vspace{-0.1cm}
\subsection{Comparison with SOTA Methods}

We compared the performance of our DSFDA method with SOTA approaches on the BioVid dataset under various experimental settings, as detailed in Table \ref{tab:SFDA-DE_result}. The \textit{Source-only model} serves as a lower-bound baseline, where the pre-trained source model is directly tested on target domain data without any adaptation. In the \textit{SFDA}, the source model is adapted to the target data using only neutral video frames from the target domain. In \textit{SFDA ($k$-shot)}, real neutral images with representative pain prototypes from the source domain are used to address the missing classes in the adaptation phase. The goal is to investigate whether pain prototypes drawn from the source domain can effectively guide the model in recognizing pain in the target domain.

As illustrated in Table \ref{tab:SFDA-DE_result}, the performance of the \textit{source-only} model, which does not utilize any adaptation, is significantly lower than the results achieved after adaptation using either synthetic pain images or pain prototypes. This highlights the importance of domain adaptation in addressing the gap between the source and target domains. As expected, the \textit{SFDA} adaptation setting, which utilizes only short neutral video frames for adaptation, significantly degrades performance, yielding even worse results than the source-only model in existing SOTA methods. When critical expression classes (pain) are missing from the target data, two SOTA methods, SHOT \cite{liang2020we}, and NRC \cite{yang2021exploiting}, experience substantial drops in accuracy, with reductions to 18.27\% and 8.31\%, respectively. These findings highlight the limitations of relying solely on neutral data for adaptation and the challenges models face when key emotional classes are unavailable during adaptation. We used SFDA methods for comparison, which closely align with our proposed method. Our proposed DSFDA method achieves 11.62\% higher accuracy than the source-only model due to the generation and adaptation of missing classes in one step through a disentanglement framework. These methods help separate identity-specific and expression-related features, allowing the model to focus on subtle variations in expressions. This leads to better generalization and improved performance, even when pain-specific data is missing. In \textit{SFDA ($77$-shot)}, we introduced source domain prototypes alongside neutral target data for adaptation, selecting representative samples from the source subjects to address the absence of pain data in the target domain. Our proposed DSFDA method outperformed both SHOT and NRC when using these prototypes, demonstrating the effectiveness of leveraging them to bridge the domain gap. Currently, we use a single image prototype for each subject from the source data. The final setting, referred to as the \textit{Oracle}, involves adapting the source model using both real neutral and pain images from each target subject. This setting serves as the upper bound for adaptation performance, providing the model full access to all relevant class information in the target domain. By utilizing both neutral and pain data during adaptation, the Oracle demonstrates the model's maximum potential when no data is missing, serving as a benchmark for comparison against more challenging settings where class information is incomplete or synthetic.

The performance of the proposed DSFDA method is compared with SOTA approaches on the UNBC-McMaster and StressID datasets under four experimental settings in Tables \ref{tab:SFDA-DE_mcmaster} and \ref{tab:stress}. DSFDA consistently outperforms the source-only baseline in both datasets. On the UNBC-McMaster dataset, our method achieves 78.75\% accuracy in the prototype setting. In the neutral-only setting, DSFDA reaches 77.70\%, outperforming \textit{SHOT} and \textit{NRC} by over 20\%.

\begin{table}[b!]
\centering
\caption{Impact of loss weighting on target accuracy on BioVid.}
\begin{tabular}{cccc|c}
\hline
$\lambda_{\text{rec}}$ & $\lambda_{\text{per}}$ & $\lambda_{\text{adv1}}$ & $\lambda_{\text{adv2}}$ & Accuracy (\%)\\ & & & & on target data \\
\hline
0 & 1 & 0.2 & 0.8 & 76.28 \\
5 & 0 & 0.2 & 0.8 & 76.50 \\
5 & 1 & 0.0 & 0.0 & 66.00 \\
5 & 1 & 0.2 & 0.8 & \textbf{81.88} \\
\hline
\end{tabular}
\label{tab:ablation-hyperparams}
\end{table}

Next, we evaluated the performance of the proposed DSFDA method and SOTA methods on the BioVid dataset for 10 target subjects using all 19 sources from UNBC-McMaster. As shown in Table~\ref{tab:mc_to_bio}, the \textit{source-only} model achieved an average accuracy of 56.48\%. Among the SFDA methods, SHOT~\cite{liang2020we, siddiqui2022fexgan} and NRC~\cite{liang2020we, wang2020imaginator} achieved average accuracy of 50.86\% and 52.30\%, respectively. Notably, the proposed DSFDA method obtained the highest average accuracy of 74.01\% and outperformed the other methods across most subjects. For more detailed information on the generation and adaptation processes within a two-stage framework, including visual results and the impact of various clustering methods on selecting source image prototypes, please refer to the supplementary material.

To better understand the limitations of our method, we conducted a qualitative analysis of failure cases. We observed that the model tends to struggle with subjects exhibiting subtle or ambiguous expressions, particularly for older subjects (e.g., Sub-10 in the BioVid dataset). In some instances, the generated images lacked fine-grained expression details that impacted the adaptation performance.

\subsection{Ablation Studies}

\noindent \textbf{Impact of hyperparameter values.}
To assess the individual contributions of each component in the generator's objective, we perform an ablation study by varying the loss weighting hyperparameters: $\lambda_{\text{rec}}$ (reconstruction loss), $\lambda_{\text{per}}$ (perceptual loss), $\lambda_{\text{adv1}}$ (identity adversarial loss), and $\lambda_{\text{adv2}}$ (expression adversarial loss). Table~\ref{tab:ablation-hyperparams} presents the results on the target domain. Disabling any of these terms by setting its coefficient to zero leads to a noticeable performance drop, which demonstrates the complementary roles of each objective. The full configuration yields the highest accuracy, validating the effectiveness of our combined loss formulation.

\begin{figure}[t!]
    \centering
    \includegraphics[width=0.49\textwidth, height=0.28\textwidth]{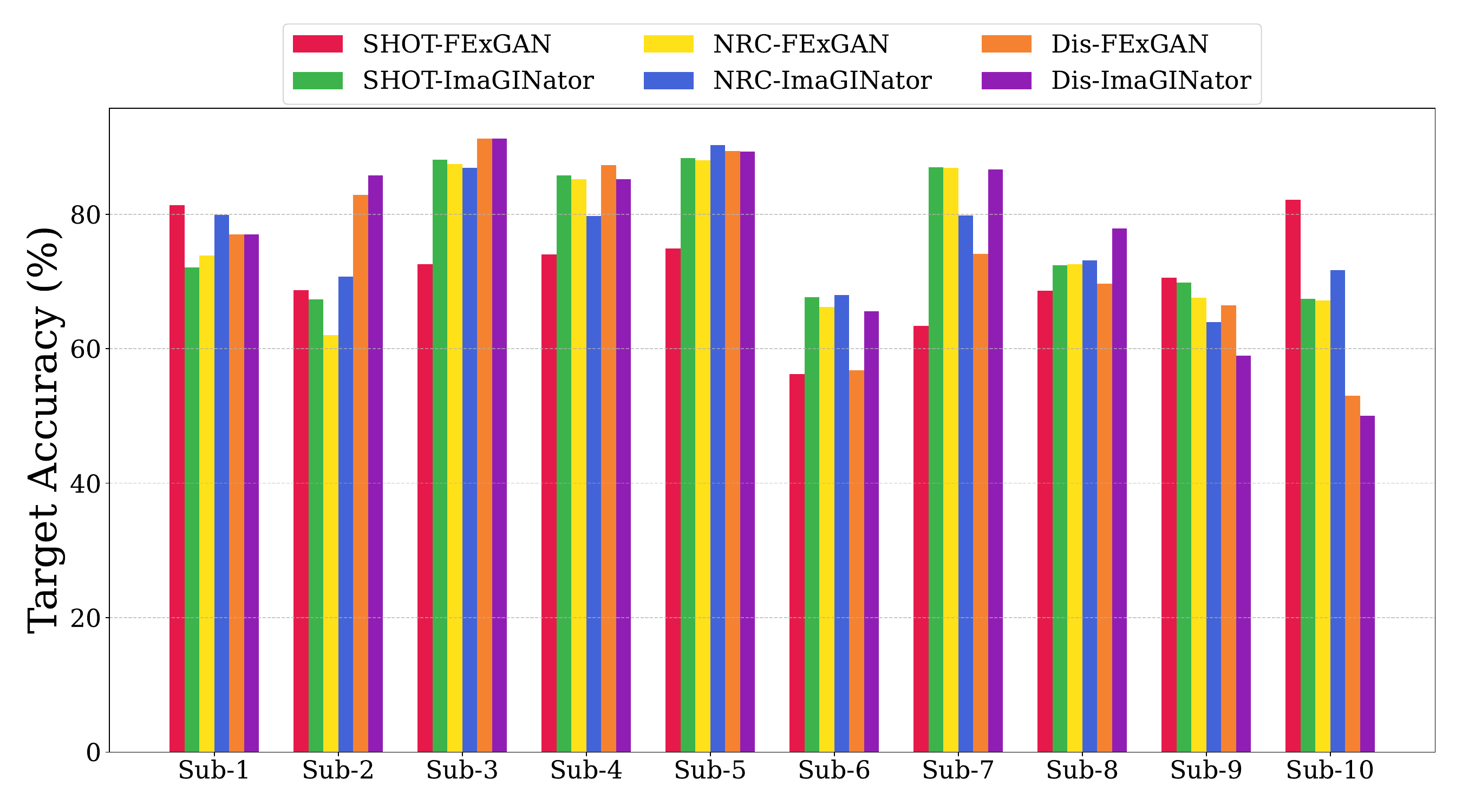}
    \caption{Per-subject target accuracies using target neutral and synthetic pain images for 10 subjects in the BioVid dataset. Each group of six bars represents the three methods (SHOT, NRC, and DSFDA) combined with two generative models (FExGAN and ImaGINator). The legend at the top identifies each method-model pairing.}
    \label{fig:gen}
\end{figure}

\noindent \textbf{Impact of different generative models.} To evaluate the impact of using different generative models (FExGAN vs. ImaGINator) for pain-image synthesis, we conduct an ablation study across three methods, SHOT, NRC, and DSFDA, on 10 target BioVid subjects. In Fig.~\ref{fig:gen}, each subject is represented on the x-axis, and the corresponding target accuracies are plotted as grouped bars. Within each group, bars reflect the chosen method paired with either FExGAN or ImaGINator. Among generative models, ImaGINator variants (e.g., SHOT-ImaGINator, NRC-ImaGINator) often surpass their FExGAN-based counterparts (SHOT-FExGAN, NRC-FExGAN), suggesting that ImaGINator generates pain expressions with greater diversity or fidelity. Furthermore, DSFDA demonstrates a consistent performance boost across most subjects, underlining the benefits of its domain-discriminative approach. Specifically, DSFDA-ImaGINator often leads to the highest accuracy, indicating strong synergy between our domain adaptation framework and a robust generative model.

\begin{figure}[t!]
    \centering
    \includegraphics[width=0.47\textwidth, height=0.25\textwidth]{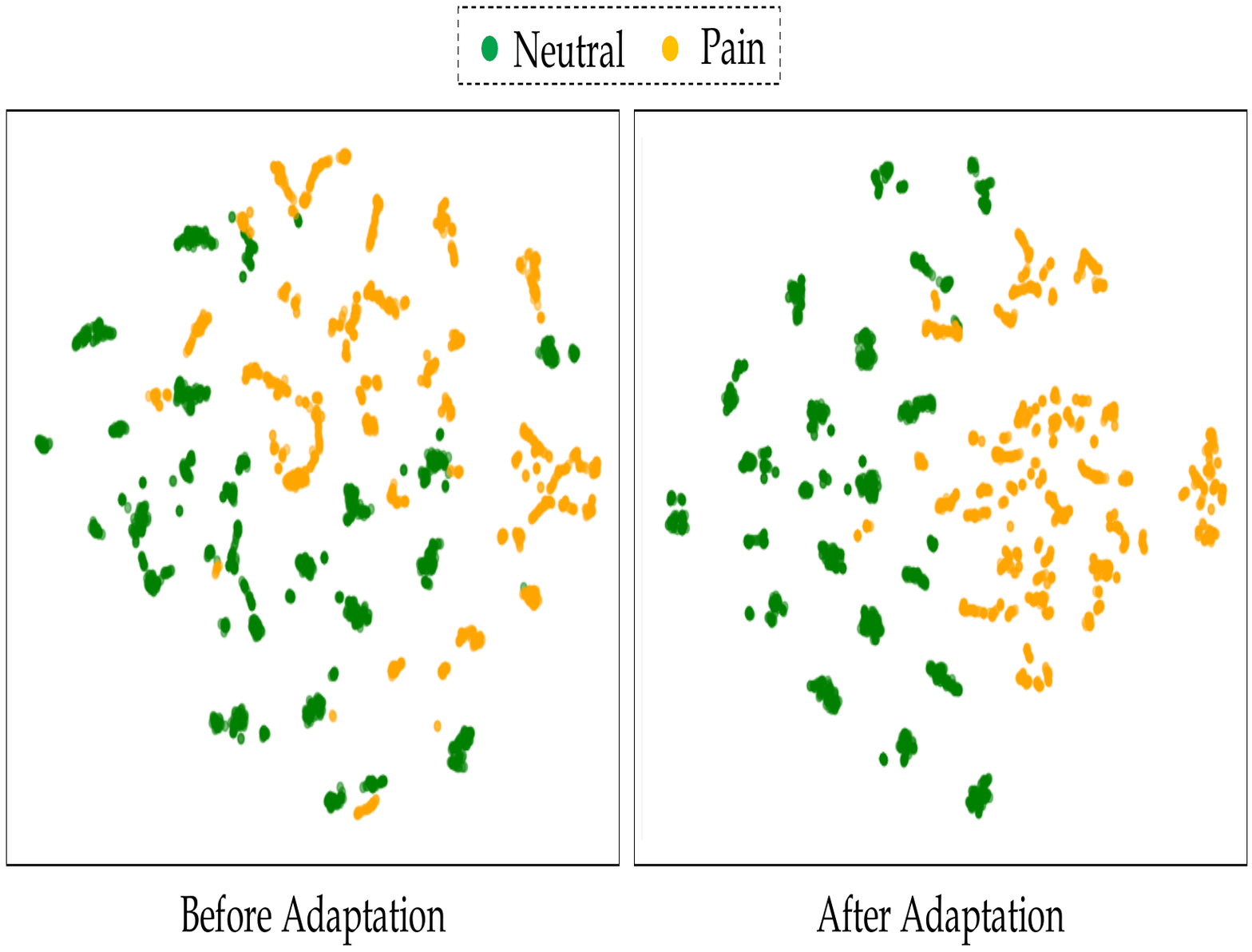}
    \caption{t-SNE visualizations of embeddings from source to target subjects for sub-1. Different colors represent different classes (green for neutral and orange for pain). (left) the source-only setting, without adaptation; (right) our DSFDA method after adaptation.}
    \label{fig:tsne}
\end{figure}

\subsection{Visualization of Feature Distributions}
Fig. \ref{fig:tsne} shows the t-SNE visualizations \cite{van2008visualizing} of the feature representations generated by the proposed DSFDA method for sub-a from the BioVid dataset. Fig.~\ref{fig:tsne}(left) displays the feature distribution before adaptation, while Fig.~\ref{fig:tsne}(right) shows the distribution after adaptation. In the source-only scenario, the features for the two classes (neutral and pain) are poorly aligned, indicating significant domain discrepancy. After applying our DSFDA method, the features are better aligned across the two classes. This improvement is due to the disentanglement strategy employed by our method, which separates domain-specific variations from class-relevant information, leading to more effective adaptation and improved alignment of feature representations.

\section{Conclusion}

This paper introduces a subject-based SFDA method for FER in real-world health diagnosis and monitoring applications from Q\&A videos. This method is specialized for the practical case where only neutral expression control target data is available for adaptation. Unlike conventional SFDA settings, our approach effectively adapts a pre-trained source model to target subjects by leveraging a disentangled generative strategy. We propose a one-stage disentangled SFDA (DSFDA) framework that integrates data generation with domain adaptation, improving model generalization while mitigating the practical constraints of FER data collection in real-world clinical environments. Experimental results on the BioVid and UNBC-McMaster pain assessment and on the StressID datasets show that DSFDA can significantly outperform SOTA SFDA methods. Results  highlight its effectiveness in handling cross-domain variability and subject-specific adaptations.\\

\noindent\textbf{Acknowledgments:} This work was supported in part by the Fonds de recherche du Quebec – Santé (FRQS), the Natural Sciences and Engineering Research Council of Canada
(NSERC), Canada Foundation for Innovation (CFI), and the Digital Research Alliance of Canada.

\section{Ethical Impact Statement}

\noindent \textbf{Potential harms to human subjects.} This study did not involve new data collection. It utilizes the publicly available BioVid Heat Pain dataset \cite{walter2013biovid} to evaluate the proposed method. Access to the BioVid dataset requires a request to the authors of the dataset. The dataset includes participants from three age groups: (1) 18--35, (2) 36--50, and (3) 51--65, with an equal distribution of male and female subjects in each group, minimizing potential age and gender bias. All participants, except for 13 subjects, provided informed consent for their data (including images and videos) to be used for research dissemination through online publications, conference presentations, and proceedings. In this paper, the data from these 13 participants were not used to illustrate research results.

Additionally, we used the UNBC-McMaster Shoulder Pain dataset \cite{lucey2011painful}, which requires submitting a request to the authors and signing an access agreement. We also employed the StressID dataset \cite{chaptoukaev2023stressid}, which includes video, audio, and physiological recordings collected under ethical approval. Access to the StressID dataset requires signing a data usage agreement, and it is limited to non-commercial scientific research. All participants provided informed consent, and data usage must comply with GDPR and privacy guidelines. Only the visual modality (video recordings) was used in this study.

\noindent \textbf {Potential negative societal impact.} The generation of pain images has the potential to be misused in order to create false evidence of pain or medical conditions in healthcare or legal settings. Pain images could be exploited to influence emotions in contexts such as fundraising campaigns or social media, where the aim may be to elicit sympathy or alarm from the audience in order to further a particular agenda.

\noindent \textbf{Limits of generalizability.} The proposed method has been validated for neutral and pain levels using the BioVid, UNBC-McMaster, and StressID datasets. However, we do not claim that the method is generalizable to all emotions or stress conditions, as further validation is necessary to confirm its effectiveness across a broader range of expressions and emotional states.

\noindent \textbf {Risk-mitigation strategies.} To mitigate the misuse risks, licensing and secure authentication protocols to restrict access to the generative model is an effective strategy. This approach ensures that only authorized users can access the models, while strict usage policies help prevent misuse. Moreover, authorities can detect and identify fake images using common and widely available tools, providing an additional layer of protection against potential misuse. While we will make our code publicly available, the pre-trained weights will not be released and will only be provided upon request from trusted parties. We will provide an ethical impact statement with the paper to mitigate the risk of misunderstanding about our method. Additionally, we will explicitly clarify which emotions this work has been validated for, emphasizing that its effectiveness beyond these validated emotions requires further investigation and validation.

\appendix

This supplementary material encompasses the following sections:

\noindent \begin{itemize}
    \item \textbf{A Two-Stage Methodology:} A detailed description of the generation and adaptation stages.
    \item \textbf{Ablation Studies:} Ablation studies on image generation methods.
\end{itemize}

\vspace{+0.4cm}
\section{Two-Stage Methodology}

\begin{figure}[!h]
    \centering
    \includegraphics[width=0.48\textwidth, height=0.26\textwidth]{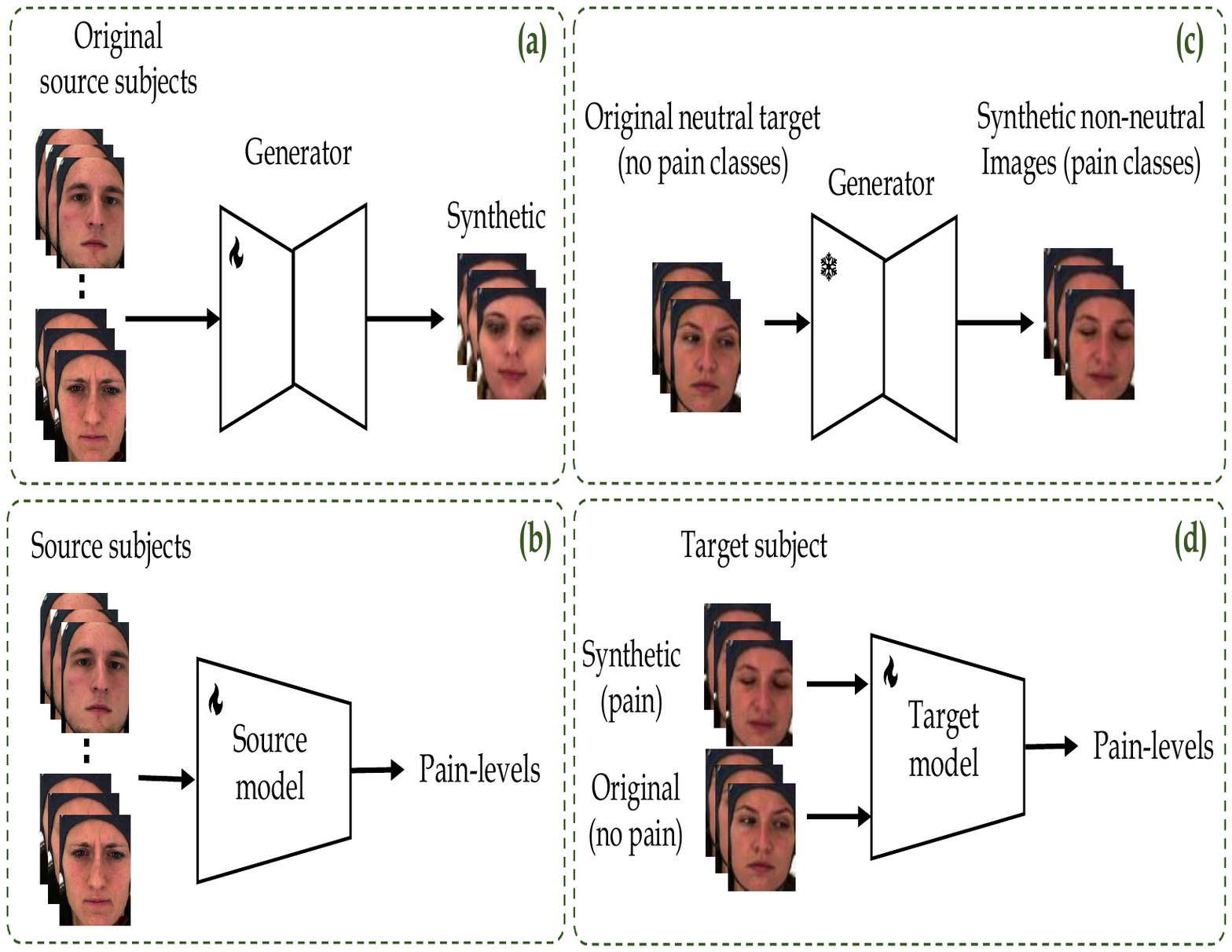}
    \caption{The two-stage method. The first stage (left): (a) the generator is trained using source subjects to generate images with different pain levels, (b) the source model is pre-trained on source subjects. Second stage (right): (c) Pain images/videos for the target subject are generated using a generative model in inference time, (d) the generated pain images/videos are combined with original neutral images for the target subject for source model adaptation.}
    \label{fig:mts}
\end{figure}

In this two-stage framework, the primary goal is to address the challenge of missing pain intensity classes in the target domain by generating video frames that represent varying levels of pain. This approach effectively converts the problem of SFDA with missing classes in the target data into a standard SFDA scenario, where the source and target domains share the same set of classes, as illustrated in Fig.~\ref{fig:mts}.

To achieve this, we leverage several advanced generative models, including VAEs, GANs, and diffusion models, to synthesize images or videos that depict a wide range of pain intensities. The research begins by thoroughly exploring different conditional GANs (cGANs) to create realistic images or videos that correspond to various pain levels. These generative models are trained on source domain data, where pain intensity labels are available, to ensure that the generated frames accurately reflect the intended pain levels.

In the first stage, Fig.~\ref{fig:mts}(left), the generator and the source model is trained on the 77 source subjects with two classes (neutral and pain). In the second stage, Fig.~\ref{fig:mts}(right), the pre-trained generator is utilized to generate missing classes (pain) for the target subject. Moreover, the pre-trained source model is used for adaptation on the real neutral and generated pain images/videos for the ten target subjects.

\begin{figure*}[t!]
    \centering
    \begin{subfigure}[b]{0.31\textwidth}
        \centering
        \includegraphics[width=\linewidth]{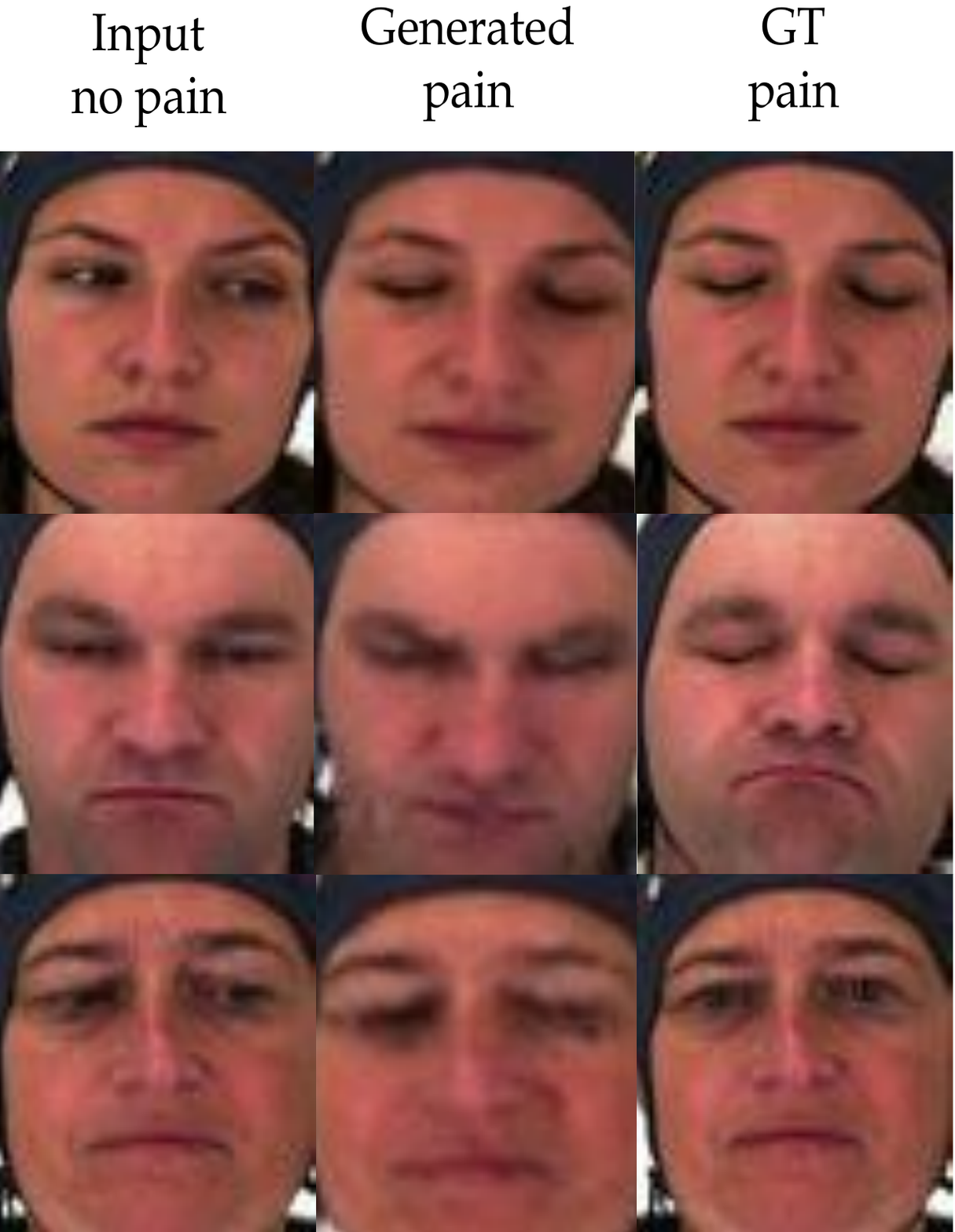}
        \caption{}
        \label{fig:ImaGINator_v}
    \end{subfigure}
    \hfill
    \begin{subfigure}[b]{0.31\textwidth}
        \centering
        \includegraphics[width=\linewidth]{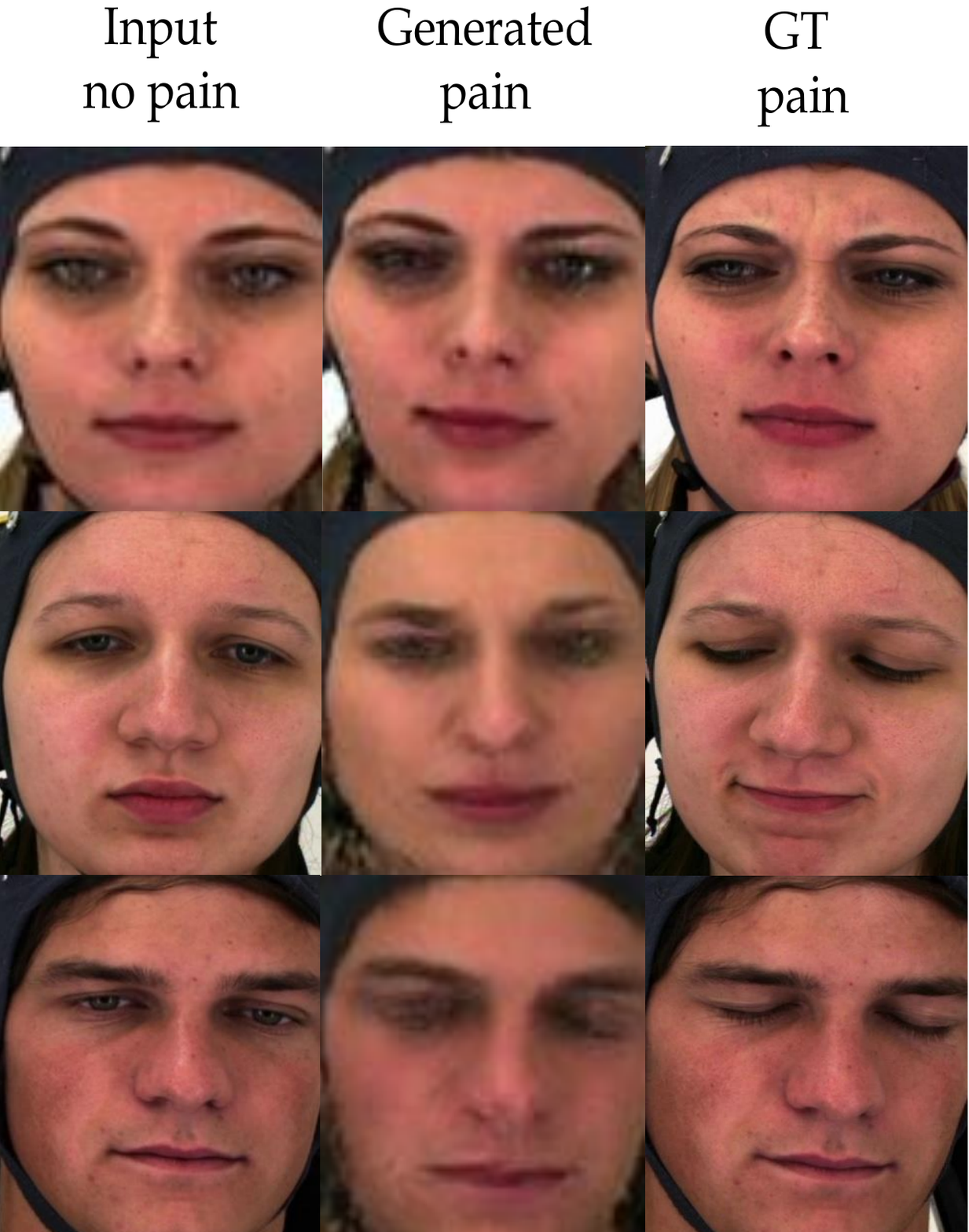}
        \caption{}
        \label{fig:FExGAN_v}
    \end{subfigure}
    \hfill
    \begin{subfigure}[b]{0.31\textwidth}
        \centering
        \includegraphics[width=\linewidth]{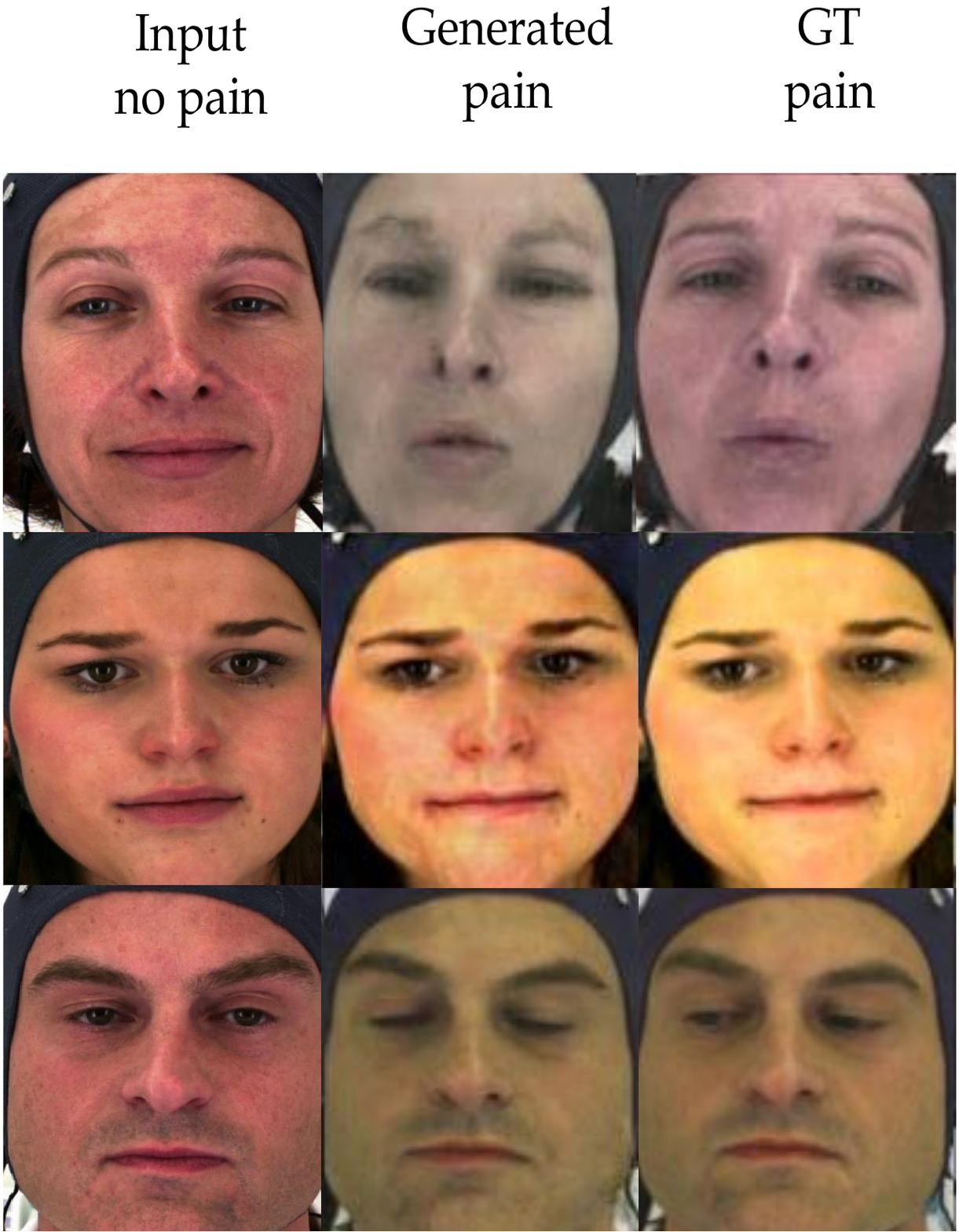}
        \caption{}
        \label{fig:Third_v}
    \end{subfigure}
   \caption{Visual comparison of generated pain images produced by different models: (a) ImaGINator, (b) FExGAN, and (c) DSFDA.}
    \label{fig:three_models}
\end{figure*}

\section{Ablation Studies}

\noindent \textbf{Ablation on the generation results.} The visual results of the generated images using ImaGINator and FExGAN-Meta are illustrated in Fig.s~\ref{fig:three_models}(a) and ~\ref{fig:three_models} (b), respectively. In addition to the visual results, we compare them using numerical results using three metrics, including structural similarity index measure (SSIM), peak signal-to-noise ratio (PSNR), and fréchet inception distance (FID). SSIM is used as a measure to evaluate the quality of generated images using generative models compared to the ground truth or target images. PSNR is used to assess how closely the generated images match the ground truth images. FID measures the similarity between the distributions of feature representations of real and generated images. In Fig.~\ref{fig:three_models} (c), the generated pain images using our DSFDA method are shown.

The quantitative performance of the generative models used to produce synthetic pain images is presented in Table \ref{tab:GANs-result}. ImaGINator achieves superior results compared to FExGAN-Meta, with higher SSIM (0.82 vs. 0.69) and PSNR (23.00 vs. 22.14), as well as a lower FID score (99.50 vs. 110.58), indicating that ImaGINator generates more realistic and high-quality images, resulting in improved adaptation performance.

We compared the results for the same settings on the UNBC-McMaster dataset using the two-stage framework in Table \ref{tab:SFDA-DE_mcmaster}. In the Source Only setting, the pre-trained source model achieves an average accuracy of 61.30\%. Using synthetic pain images in the Org. Neutral \& Syn. Pain setting improves the results of adaptation significantly compared to the Source Only setting. In this category, SHOT (68.27\%) and NRC (68.37\%) show comparable results, while the proposed method, SFDA-DE (ours), outperforms both with an average accuracy of 70.41\%, demonstrating the effectiveness of our approach. Finally, in the Oracle setting, where the model is fine-tuned on labeled target data, the framework achieves the highest overall accuracy of 97.59\%, setting the performance ceiling under ideal conditions.

\begin{table}[b!]
  \centering
  \caption{Quantitative results of the generative models on BioVid dataset.}
  \label{tab:GANs-result}
  \begin{tabular}{|l |c|c|c|}
    \hline
    Method & SSIM $\uparrow$ & PSNR $\uparrow$ & FID $\downarrow$ \\[3pt]
    \hline
    \hline
    ImaGINator & \textbf{0.82} & \textbf{23.00} & \textbf{99.50} \\[3pt]
    \hline
    FExGAN-Meta & 0.69 & 22.14 & 110.58 \\[3pt]
    \hline
  \end{tabular}%

\end{table}

\begin{figure}[H]
    \centering
    \includegraphics[width=0.5\textwidth, height=0.25\textwidth]{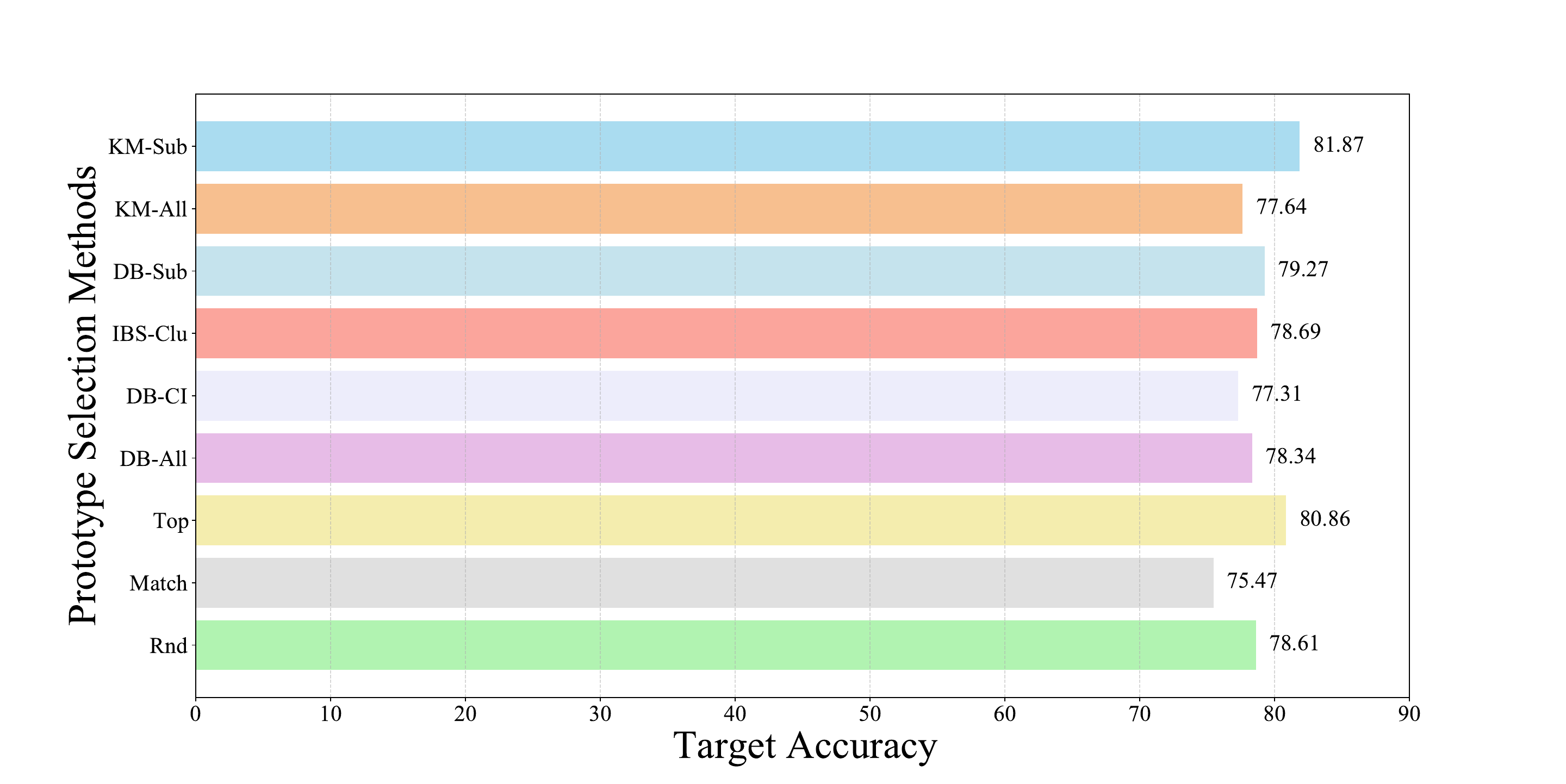}
    \caption{Performance comparison of various prototype selection methods applied to the source data. Methods include random selection (Rnd), direct matching (Match), correctly classified images (Top), DBSCAN clustering across all subjects (DB-All), $K$-means clustering across all subjects (KM-All), DBSCAN clustering per subject (DB-Sub), $K$-means clustering per subject (KM-Sub), DBSCAN with the densest cluster (DB-D), and DBSCAN with the closest cluster to the target subject (DB-Cls).}
    \label{fig:pro-methods}
\end{figure}

 \noindent \textbf{Impact of prototype selection methods.} We evaluated the effectiveness of various image prototype selection strategies from the source domain by comparing random, match, DBSCAN, and $k$-means clustering methods and their impact on adaptation performance, as shown in Fig.~\ref{fig:pro-methods}. For DBSCAN, we varied the hyperparameters epsilon (the maximum distance between two points for them to be considered neighbors) and min\_samples (the minimum number of points required to form a dense region). These changes result in different clustering outcomes. In DB-All and DB-Sub, the algorithm was applied to all 77 subjects collectively and each subject individually. In DB-D, the most dense cluster for each subject was selected, while in DB-Cls, we selected the cluster closest to the target subject's cluster. For the $K$-means we used two configurations: $K$=$77$ (KM-All), where the closest point to the centroid was chosen for all subjects, and $K$=$1$ 

    \noindent (KM-Sub), where the closest point to the centroid was selected for each subject individually. Based on the results, subject-specific clustering significantly improved performance in both cases. Applying DBSCAN to individual subjects outperformed the all-subject approach by 3.72\%, while for $K$-means, the improvement was 4.83\%. Based on the average results, utilizing matched videos from the source domain significantly enhances performance more than using a random video. This improvement can be attributed to the shift in expression between different subjects; each individual expresses pain in distinct ways and with varying intensity. For instance, facial cues may be more pronounced in women and younger subjects compared to men and older subjects. Consequently, the model can better disentangle expression from other information when using videos that align with the age and gender of the target subjects.

\begin{table}[t!]
\centering
\LARGE
\renewcommand{\arraystretch}{1.5} 
\caption{Accuracy (\%) of the proposed method and state-of-the-art methods on the UNBC-McMaster dataset for 5 target subjects with all 20 sources. Bold text shows the highest accuracy.}\label{tab:SFDA-DE_mcmaster}
\resizebox{0.47\textwidth}{!}{%
	\begin{tabular}{c|c|ccccc|c}
		\hline
		\textbf{Setting} & \textbf{Methods} & \textbf{Sub-1} & \textbf{Sub-2} & \textbf{Sub-3} & \textbf{Sub-4} & \textbf{Sub-5} & \textbf{Avg.} \\
	    \hline
     \hline
         Source only & Source model & 68.56 & 78.32 & 60.37 & 65.54 & 33.72 & \textbf{61.30} \\
        \hline
        \multirow{3}{*}{\makecell{Org. neutral \\\& syn. pain}} & SHOT \cite{liang2020we} & 78.24 & 84.39 & 60.37 & 77.19 & 41.18 & \textbf{68.27} \\
        & NRC \cite{yang2021exploiting} & 78.24 & 84.39 & 61.00 & 77.05 & 41.18 & \textbf{68.37}\\
        & SFDA-DE (ours) & 78.73 & 84.39 & 67.88 & 79.00 & 42.05 & \textbf{70.41}\\
        \hline
        Oracle
        & Fine-tune & 95.63 & 98.44 & 98.11 & 97.54 & 98.25 & \textbf{97.59}\\
        \hline
	\end{tabular}
}
\end{table}
\begin{figure}[t!]
   \centering
   \includegraphics[width=0.47\textwidth, height=0.28\textwidth]{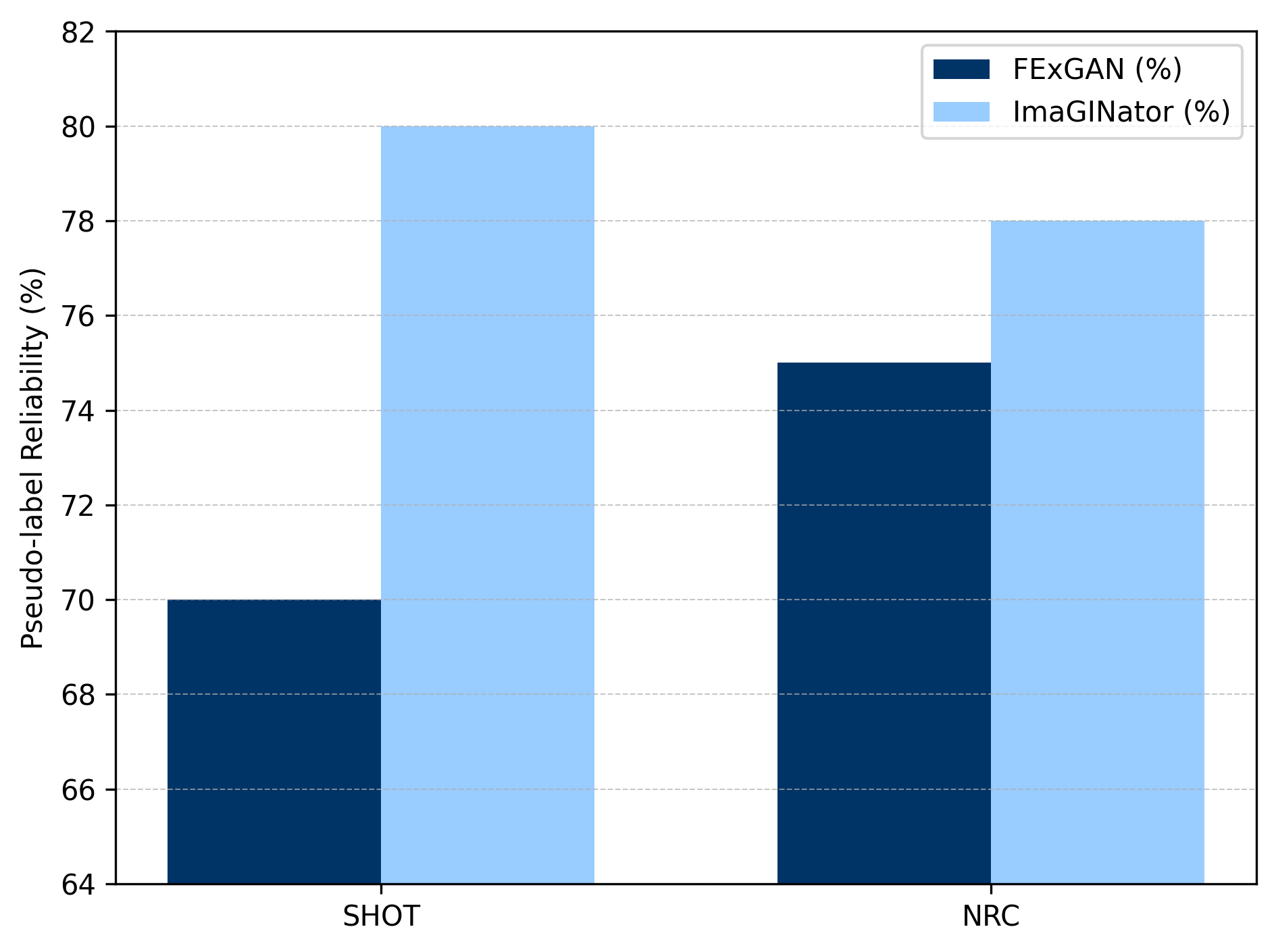}
   \caption{Comparison of pseudo-label generation reliability between SHOT and NRC methods using FExGAN and ImaGINator to generate missing classes.}
    \label{fig:pseudo-label}
\end{figure}

    \noindent \textbf{Impact of generation methods on pseudo-labels.} We evaluated the reliability of pseudo-labels produced by SHOT and NRC methods when using FExGAN and ImaGINator to generate the missing classes in the target data. ImaGINator consistently produced a higher ratio of reliable pseudo-labels compared to FExGAN. Specifically, the NRC method achieved 75\% reliability with FExGAN and improved to 78\% with ImaGINator, while the SHOT method resulted in 70\% with FExGAN and reached 80\% with ImaGINator. Notably, the combination of SHOT with FExGAN yielded the lowest pseudo-label ratio (70\%), suggesting that SHOT is less effective when paired with FExGAN compared to ImaGINator. Furthermore, the NRC-FExGAN configuration outperformed our DSFDA-FExGAN method, likely because the generated pain images -- often resembling neutral expressions -- received reliable pseudo-labels. Since most real or generated pain images do not visibly express pain, assigning pain labels indiscriminately can introduce misleading representations, ultimately diminishing the model's performance during testing.


\balance
{\small
\bibliographystyle{IEEEtran}
\bibliography{egbib}
}
\vfill
\end{document}